\documentclass[sigconf]{acmart}

\usepackage{booktabs} 

\usepackage{caption} 
\usepackage[justification=centering]{caption}

\usepackage{graphicx}
\graphicspath{ {images/} }
\usepackage{subcaption}

\usepackage{tabu}
\usepackage{booktabs}

\setcopyright{rightsretained}

\acmConference[GECCO '18]{the Genetic and Evolutionary Computation Conference 2018}{July 15--19, 2018}{Kyoto, Japan}
\acmYear{2018}
\copyrightyear{2018}

\begin{document}
\newcommand{\comm}[1]{}

\title{Why Don't the Modules Dominate?}
\subtitle{Investigating the Structure of a Well-Known Modularity-Inducing Problem Domain} 

\author{Zhenyue Qin}
\affiliation{%
  \institution{Australian National University}
  \streetaddress{Building 108}
  \city{Canberra}
  \state{ACT}
  \postcode{2601}
  \country{Australia}
}
\email{zhenyue.qin@anu.edu.au}

\author{Tom Gedeon}
\affiliation{%
  \institution{Australian National University}
  \streetaddress{Building 108}
  \city{Canberra}
  \state{ACT}
  \postcode{2601}
  \country{Australia}
}
\email{tom.gedeon@anu.edu.au}

\author{R I (Bob) McKay}
\affiliation{%
  \institution{Australian National University}
  \streetaddress{Building 108}
  \city{Canberra}
  \state{ACT}
  \postcode{2601}
  \country{Australia}
}
\email{robert.mckay@anu.edu.au}

\renewcommand{\shortauthors}{Qin et al.}

\begin{abstract}
Wagner's modularity inducing problem domain is a key contribution to the study of the evolution of modularity, including both evolutionary theory and evolutionary computation. We study its behavior under classical genetic algorithms. Unlike what we seem to observe in nature, the emergence of modularity is highly conditional and dependent, for example, on the eagerness of search. In nature, modular solutions generally dominate populations, whereas in this domain, modularity, when it emerges, is a relatively rare variant. Emergence of modularity depends heavily on random fluctuations in the fitness function; with a randomly varied but unchanging fitness function, modularity evolved far more rarely. Interestingly, high-fitness non-modular solutions could frequently be converted into even-higher-fitness modular solutions by manually removing all inter-module edges. Despite careful exploration, we do not yet have a full explanation of why the genetic algorithm was unable to find these better solutions.
\end{abstract}

\ccsdesc[500]{Applied computing~Biological networks}
\ccsdesc[300]{Computing methodologies~Artificial intelligence}
\ccsdesc[300]{Software and its engineering~Abstraction, modeling and modularity}

\keywords{Biological networks, Evolution of modularity}

\maketitle

\section{Introduction}
Adaptability is an essential property of both biological and artificial evolutionary systems~\cite{yang2013swarm}. 
\comm{In other words, what can we do to facilitate engineered robots to evolve in order to adapt themselves to the constantly changing environments, just like biological organisms}
Biological organisms have already solved this problem through evolutionary adaptations, of which  modularity is believed crucially important~\cite{Gerhart8582}, providing hope for artificial evolutionary systems to also generate modular, adaptable systems~\cite{pfeifer2001understanding}.
In artificial systems, modularity is also desirable for comprehensibility/engineering reasons, a good example being the "high cohesion, low coupling" principle in software engineering~\cite{coad1991object}. This emphasis on modularity was key to the software engineering boom over the last few decades~\cite{hitz1995measuring}. 

Lack of modularity has been shown as a key limitation of artificial biological systems in scaling to higher complexity~\cite{kashtan2005spontaneous,pfeifer2006body}. Artificial neural networks are usually densely connected~\cite{jain1996artificial}, whereas human brains have modular components taking different responsibilities, such as the hippocampus for novel situations and amygdala for emotional controls~\cite{coward2013towards}. Thus it is important to understand the conditions leading to the spontaneous biological emergence of modularity. Engineers may leverage them to generate modular systems, able to solve more complex problems and to autonomously adapt to new environments. Conversely, such understanding can help to winnow the plethora of evolutionary theories in biology.

Formally, modularity is the divisibility of structures or functions into sub-units that perform autonomously~\cite{schlosser2004modularity}. 
Thus a module is a group of elements which associate preferentially within the group~\cite{newman2006modularity,espinosa2010specialization}. 
Many biological activities and structures can be modeled in the form of networks -- animal brains, signaling pathways, etc.~\cite{barabasi2004network}. A network is modular if it can be partitioned into highly connected components, and between these components there are only sparse connections~\cite{freeman1977set,clune2013evolutionary}. 
Elements within a module will tend to undertake coherent functions independent of outside elements~\cite{espinosa2010specialization,larson2016recombination}. 
Such modules appear everywhere in biology~\cite{coward2013towards}, at multiple levels of biological organization~\cite{espinosa2010specialization,coward2013towards}. 
Modularity can promote the evolvability of organisms, defined as the ability to rapidly adapt to novel environments~\cite{pigliucci2008evolvability}. Modular networks allow changes in one module without disturbing other modules; and modular structures can be reused and recombined to perform new functions~\cite{espinosa2010specialization,wagner1996perspective}. 

Despite decades-long research interest in modularity~\cite{wagner2007road}, there is no consensus on its biological origin~\cite{wagner2004role,espinosa2010specialization}. 
\comm{please take extreme care to avoid sounding teleological in _any_ evolutionary/EC writing:
'evolutionary direction in biology '}
Two theories stand out, because their preconditions commonly arise in nature~\cite{wagner2007road}: modularly-varying evolutionary goals~\cite{kashtan2005spontaneous} and specializations in gene activity patterns~\cite{espinosa2010specialization}. In the former, modular changes in environments may generate an impetus toward modularity~\cite{kashtan2005spontaneous}. Organisms whose environmental sub-components change repeatedly show more modularity than those from stable environments~\cite{parter2007environmental}. Fluctuations are omnipresent in real environments~\cite{espinosa2010specialization,yachi1999biodiversity}. However it is unclear to what extent these fluctuations are modular~\cite{espinosa2010specialization}. 

Espinosa-Soto and Wagner\footnote{Since we repeatedly need to refer to this crucial paper, we abbreviate it as ES\&W} studied the emergence of modularity in an artificial environment requiring gene specialization~\cite{espinosa2010specialization}, in which gene regulatory networks need to be able to regulate toward multiple different patterns. This is a common occurrence in nature: the same collection of genes frequently exhibits different activity patterns at different phases of development or in different locations in an organism~\cite{jones1980cellular}. In ES\&W's model, the network was initially encouraged to regulate toward a single target. As evolution proceeded, further targets were added. The targets were modularly structured, sub-components changing regularly but modularly. They argued that too many interconnections between subcomponents of a network following independently changing targets would hamper the ability of each sub-component to follow its own target. They were able to demonstrate the emergence of modularity under such conditions, using a recombination-free evolutionary algorithm. The results were persuasive, but also exhibited one key difference from nature: although modular structures emerged, they did not dominate populations, whereas in biology, once modular structures emerge, they become universal across a species. Larson et al.~\cite{larson2016recombination} subsequently extended this work (within a modified environment) to explore the effects of recombination on modularity emergence. 
\comm{conditions under which gene regulatory networks started exhibiting modular structures  They concluded that modularity could arise as a by-product of gene specializations  Specifically, the distinct sub-components in the regulatory network to regulate sharing and different gene activity patterns will hamper each other's performance. Thus, modular networks that favor fewer connections between modules of the network will break the pleiotropic effect of regulating sharing and distinct gene activity patterns. Moreover, additional gene activity patterns can further improve the modularity. Their work is persuasive since the phenomena that gene regulatory networks acquire new gene activity patterns is ubiquitous in evolution.  Their theory can also act as an alternative explanation of why modular-varying environments result in modularity since organisms need to express different gene patterns for different environments~\cite{kashtan2005spontaneous,espinosa2010specialization}. However, the experiments of ES\&W lacked the crossover phase in their evolutionary simulations. Biologically, crossover is necessary.}

Our original intention was to further extend this stream, particularly emphasizing the effects of two ubiquitous biological phenomena, diploidy and crossover. To establish a baseline, we initially experimented with variants of standard genetic algorithms, yielding anomalous and difficult-to-explain results. Until these are fully explored, our original goal had to be postponed, as we could not reliably explain any results that might be obtained. The remainder of this paper details our deeper exploration of the fitness landscape of this intriguing environment. Some understanding has emerged, though we cannot yet say that we have fully understood the environment, and some paradoxical results remain. We do not see this as a criticism of the original use of the model, as it was eminently successful in challenging the view that cyclic repetition of modularly structured environments was essential for the emergence of modularity. It is its extended use as a test-bed for understanding modularity that is at issue here: whether further investigation of the comparative behavior of different algorithms on this problem gives any real insight into the equivalent behavior either in biology, or in real-world application of evolutionary algorithms.

\comm{In this paper, we aim to investigate the plausibility of the theory stating that gene specialization drives modularity of organisms~\cite{espinosa2010specialization}. We will first explore whether there exist methods that can expedite the evolutionary process. For example, crossover is assumed to be an effective method to enhance the efficacy of combining useful traits in evolutionary simulations. Therefore, it is beneficial to explore whether there exists a crossover mechanism that can promote modularity. Similarly, the elitism, which is another common mechanism utilized in the artificial evolution, is also worthwhile exploring its contribution to the computational evolution. Furthermore, we will also investigate whether different fitness evaluation methods give rise to different modularity levels. 

Moreover, experiments in~\cite{espinosa2010specialization} did not demonstrate whether structures with high modularity has gained a dominant status on survivability. In biology, there is no organism that exhibits non-modular structures. As such, it is plausible to assume non-modular creatures have been extinct from history. Therefore, modular individuals are expected to have far better performance than non-modular ones, especially for complicated environments. As such, it will be constructive to investigate the dominant status of modularity on survivability by comparing the fitness values of the eminent modular organisms to less modular ones.}

\comm{This sounds way too teleological: Furthermore, we also wish to discover what properties of modular structures will obtain in a long-term evolution. That is, towards what direction is the system with high modularity evolving? Although the experiments suggested a significant emergence of modular structures was due to gene specialization, they only reveal specialization is the origin of modular structures. It did not explain the evolutionary direction of modular systems. }

\section{Methods}
We use genetic algorithms as our evolutionary simulation tools. The gene regulatory network that we used in this paper was originally proposed by Wagner~\cite{wagner1996does} and customized by ES\&W~\cite{espinosa2010specialization} and Larson et al.~\cite{larson2016recombination}. 
All simulation code was implemented in Java 1.8.0 and Python 2.7.10. They are all publicly available at 
\begin{anonsuppress}
  https://gitlab.cecs.anu.edu.au/u5505995/Modularity-Exploration.git. 
\end{anonsuppress}
Modularity was evaluated using the NetworkX package with the community API~\cite{hagberg2008exploring}. 
All the generated data can be downloaded at: 
\begin{anonsuppress}
  goo.gl/9PF4m7
\end{anonsuppress}

\subsection{Model}
Cells in an organism display heterogeneity in functionalities and morphologies, yet generally contain the same set of genes. 
\comm{This isn't strictly true. Sperm and eggs don't have the same genes as other cells (since they are haploid), some immune cells evolve rapidly within an individual, but more generally, many cell lines evolve slowly during lifetime - or rapidly if cancer occurs; there are probably other exceptions we don't know yet}
Cells interpret the same genetic material in different ways so that their behaviors and structures vary. These distinct interpretations are due to regulation, among other mechanisms via the activation and repression of genes by other genes~\cite{wagner1996does}: the effects of different genes are not mutually independent. A protein that is generated by one gene may activate or repress another. 
This mechanism can be usefully abstracted by a weighted directed graph. In this graph, absence of an edge denotes lack of interaction, while a further abstraction limits the weights to +1 (activation) or -1 (repression)~\cite{wagner1996does}. The term \textquotedblleft gene activity pattern\textquotedblright\ describes the activeness status of the entire set of genes. Different activity patterns generally imply distinct cellular functions and forms~\cite{espinosa2010specialization}. 

We used ES\&W's~\cite{espinosa2010specialization} representation of a gene regulatory network. The genotype of a gene regulatory network with $N$ genes is represented as an $N^2$ adjacency matrix $A = a_{ji}$. Each entry $a_{ji}$ is restricted to be either 1, 0 or -1, represents activation, independence or repression of gene $i$ by gene $j$. The gene activity pattern of this network at time $t$ is a Boolean row vector $s_{t} = [s_{t}^0,...,s_{t}^{N-1}]$. Gene $i$ can either be active ($s_t^i=1$) or inactive ($s_t^i=-1$). The state transition is modeled by:
\begin{equation}
s_{t+1}=\sigma[\sum_{j=1}^{N}a_{ji}s_t^j]
\end{equation}
where $\sigma(x)$ equals 1 if $x>0$ and is -1 otherwise. 
\comm{We have already said that the state is either -1 or 1, so it can't be 0}

\subsection{Fitness}
A common role for GRNs is to maintain specific activation states in cells in the face of random external perturbations~\cite{aderems2005systems}. ES\&W abstracted this by defining a sequence of two (or more) target states. 

The ability of a GRN to robustly maintain a specific state was measured by randomly generating a set of $P$ perturbations of the target, with each gene in the target having a 0.15 probability of being mutated to the opposite state (ES\&W used $P=500$, and Larson et al. $P=300$). To each perturbation, the GRN was recursively applied. Preliminary experiments indicated that it normally took fewer than 20 transitions to reach an attractor~\cite{wagner1996does}. If the GRN reached a stable attractor in fewer than 20 GRN steps from the perturbation, the Hamming Distance $D$ between the attractor and the target state was returned; if 20 steps was insufficient, the maximum possible Hamming distance $D_{max}$ was returned. In either case, the value $\gamma_i=(1-D/D_{max})^5$ was computed for each perturbation $i$, with $1 \leq i \leq P$. Finally, the mean value $\bar{\gamma}$ over all $\gamma_i$ was used to compute the fitness of the GRN $g$ over a particular target $t$as:
\begin{equation}
  f_t(g)=1-e^{-3\bar{\gamma}}
\end{equation}
This process, of randomly sampling a set of perturbations of the target, and evaluating the GRN's ability to robustly return them to the target, was repeated each generation.

In the first stage, the system was evolved to regulate the first target state alone. In subsequent stages, the fitness function rewarded regulation of newly introduced states, while maintaining pressure to regulate earlier states, by computing the overall fitness $f(g)$ as the arithmetic mean of $f_t(g)$ over all targets.

\comm{Thus, non-stable attractors are assumed to be those gene regulatory networks that take more than 20 steps to attain the stability, or are cyclically stable. They are treated to have a maximum Hamming distance $D_{max}$.That is, a successful network is able to regulate a corrupted pattern to its initial form. Then,  original pattern was calculated.  This is followed by a calculation of the contribution from each perturbation attractor to the fitness, which is defined as a developmental trajectory $\gamma=(1-D/D_{max})^5$~\cite{espinosa2010specialization}. Afterwards, this process is repeated to determine 75 $\gamma_{i}$, . Finally, the fitness of a network is calculated as
\begin{equation}
f(g)=1-e^{-3g}
\end{equation}
where $g$ represents the arithmetic mean of the sum of all $\gamma_{i}$~\cite{espinosa2010specialization}. As to cases where there are more than one gene activity patterns, the arithmetic mean of $f(g)$ for all the patterns was take. Consequently, a gene regulatory network with a high fitness is able to lead to different attractors matching different targets. }

We followed this strategy, using only two targets: evolving for 500 generations with a single target, then introducing the second target for a further 1500 generations. We based our choice of the number of perturbations (75) on a trade-off between Totten's observation~\cite{totten2015exploring} that 75--100 perturbations are sufficient for emergence of modularity, and the practical need to minimize runtime.

\comm{The fitness here evaluates the likelihood that an attractor is obtained when facing perturbations~\cite{espinosa2010specialization}. In other words, ES\&W imposed a bias of robustness on their gene regulatory network models in order to indirectly select modular networks. This is because modular networks can limit perturbations in a module so that the overall structure will not be heavily affected . That is, more modular networks are more robust.}

\comm{There are two or more stages in their experiments on discovering the conditions under which modularity starts emerging. In the first stage, gene regulatory networks are evolved under selective pressure towards regulating a particular gene activity pattern, while facing some perturbations. The original gene activity pattern before perturbation is called a target. In the second and further stages, networks are evolved under selective pressure to regulate new gene activity patterns, while preserving the ability to regulate the old patterns. In the particular case where there were two gene activity patterns, the first stage lasted for 500 generations and the second took another 1500 generations.}

\comm{The perturbations of targets are randomly generated in every generation when evaluating the fitness of gene regulatory networks. In ES\&W's experiments, a network would face 500 perturbations comprising different corrupted versions of gene activity patterns. Each gene will have a probability of 0.15 to be perturbed into its opposite activity. A further study was conducted to explore a sufficient number of perturbations in order to shorten the computational time while maintaining a similar eventual improved modularity. It was concluded that 75 or 100 perturbations would lead to the noteworthy emergence of modularity~\cite{totten2015exploring}. Therefore, 75 perturbations are undertaken for evaluating the fitness of each gene regulatory network in order to reduce the running time.}

Larson et al. applied a different approach to evaluating the fitness of networks~\cite{larson2016recombination}. Instead of sampling a new set of perturbations each generation, they instead sampled a static (but larger) set of perturbations at the beginning of each run, and utilized this same set of corrupted targets whenever network fitness was calculated. This method has important computational cost advantages, since the fitness value of a given GRN on a given target remains fixed from generation to generation, so that caching and hashing methods can be used to give substantial speedups. However, it converts the original stochastically dynamic fitness evaluation into a static, deterministic one. ES\&W's fitness landscape fluctuates stochastically each generation, whereas Larson et al.'s remains fixed. This has potential implications for search.

\comm{the evolutionary landscape of individuals under this fitness evaluation will remain unchanged in each generation. On contrast, ES\&W's fitness evaluation will lead to the evolutionary landscape to shift every generation. }

\subsection{Evolutionary Simulations}
\label{subsec:bias}
ES\&W imposed a mutation bias towards networks with a specific, relatively low, link density~\cite{espinosa2010specialization}. A node in the network has a probability $\mu=0.05$ to mutate every generation; if it mutates, it can either lose or gain an interaction. The probability for a node to lose an interaction is defined as
\begin{equation}
p(u)=\frac{4r_{u}}{4r_{u} + N - r_{u}}
\end{equation}
where $N$ is the number of nodes in the network, and $r_{u}$ is the number of regulators of gene $u$~\cite{espinosa2010specialization} -- that is, the number of genes that exert effects on gene $u$. Complementarily, the probability for a gene $u$ to gain an interaction is defined to be $1-p(u)$. This bias acts to preserve the sparseness of the network, which computational biology research suggests is necessary for modularity to emerge. 

\begin{figure}[htb]
	\centering
	\begin{subfigure}[b]{0.45\linewidth}
		\includegraphics[width=\linewidth]{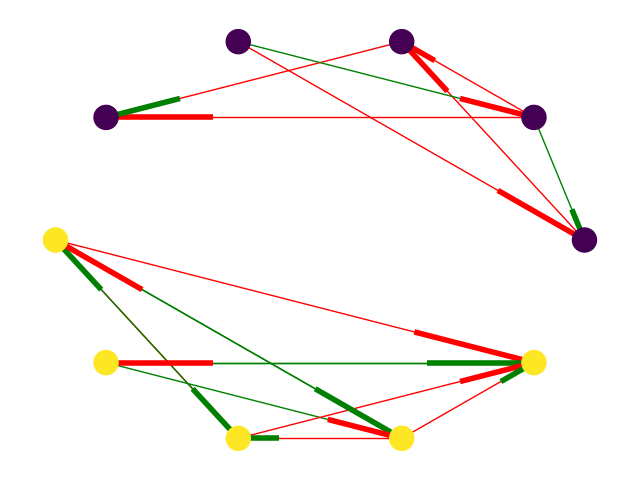}
		\caption{A modular example}
	\end{subfigure}
	\begin{subfigure}[b]{0.45\linewidth}
		\includegraphics[width=\linewidth]{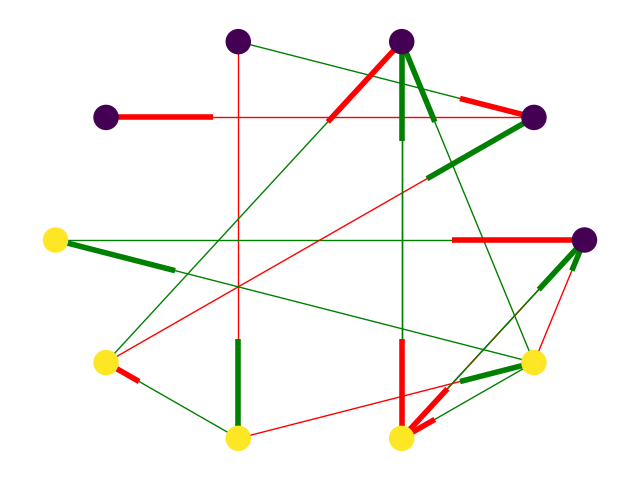}
		\caption{A non-modular example}
	\end{subfigure}
	\caption{Modular and Non-modular networks. Different node colors represent distinct modules (based on the modular changes in target values). Green and red edges mean activation and repression.}
	\label{fig:network-example}
\end{figure}
ES\&W defined modules in terms of the components of targets that followed similar activation pattern histories. In the most-used example, which used two targets of length ten, the activations of the first five locations in both targets were identical, while the activations of the second five were inverted between the targets. Thus the modules were assumed to be the connected components in the GRN involving nodes 1--5 and nodes 6--10.
Figure \ref{fig:network-example} illustrates typical examples of modular and non-modular networks.\footnote{These color conventions are used throughout this paper. While color in the images conveys additional information, the key distinctions are still observable in black and white.}

ES\&W did not use a crossover mechanism in their simulation~\cite{espinosa2010specialization}. In the reconstructed model of Larson et al., they limited crossover to nine possible partition locations of a 10-node network, corresponding to nine possible rows for splitting the adjacency matrix of a network horizontally~\cite{larson2016recombination}. We call this horizontal crossover. When two matrices $A_{1}$ and $A_{2}$ are selected for crossover at index $i$, matrices of their children will be produced as \\ \\
$C_{1}[0: i-1, :] = A_{1}[0: i-1, :]\\
C_{1}[i: 9, :] = A_{2}[i: 9, :]\\
C_{2}[0: i-1, :] = A_{2}[0: i-1, :]\\
C_{2}[i: 9, :] = A_{1}[i: 9, :]\\$

\begin{figure}[htb]
	\centering
	\begin{subfigure}[b]{0.4\linewidth}
		\includegraphics[width=\linewidth]{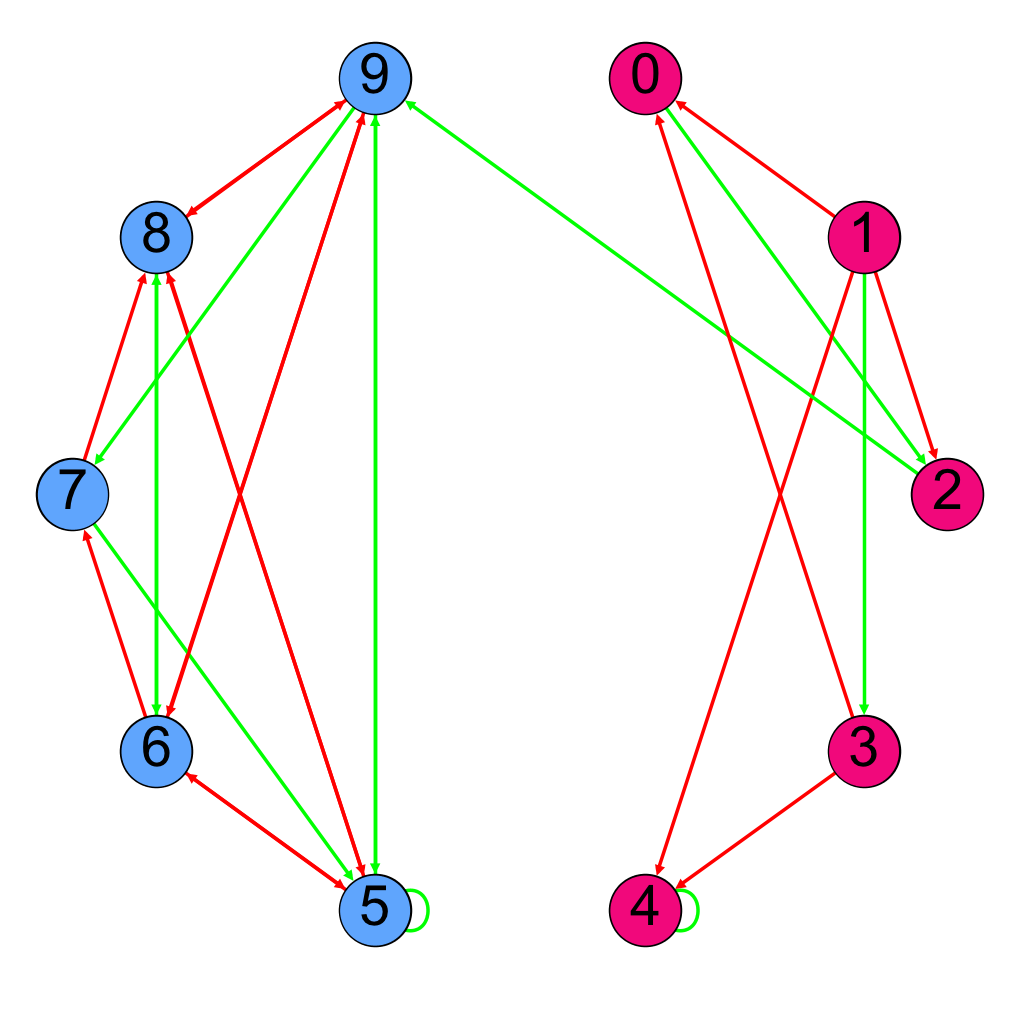}
		\caption{Parental network 1.}
	\end{subfigure}
	\begin{subfigure}[b]{0.4\linewidth}
		\includegraphics[width=\linewidth]{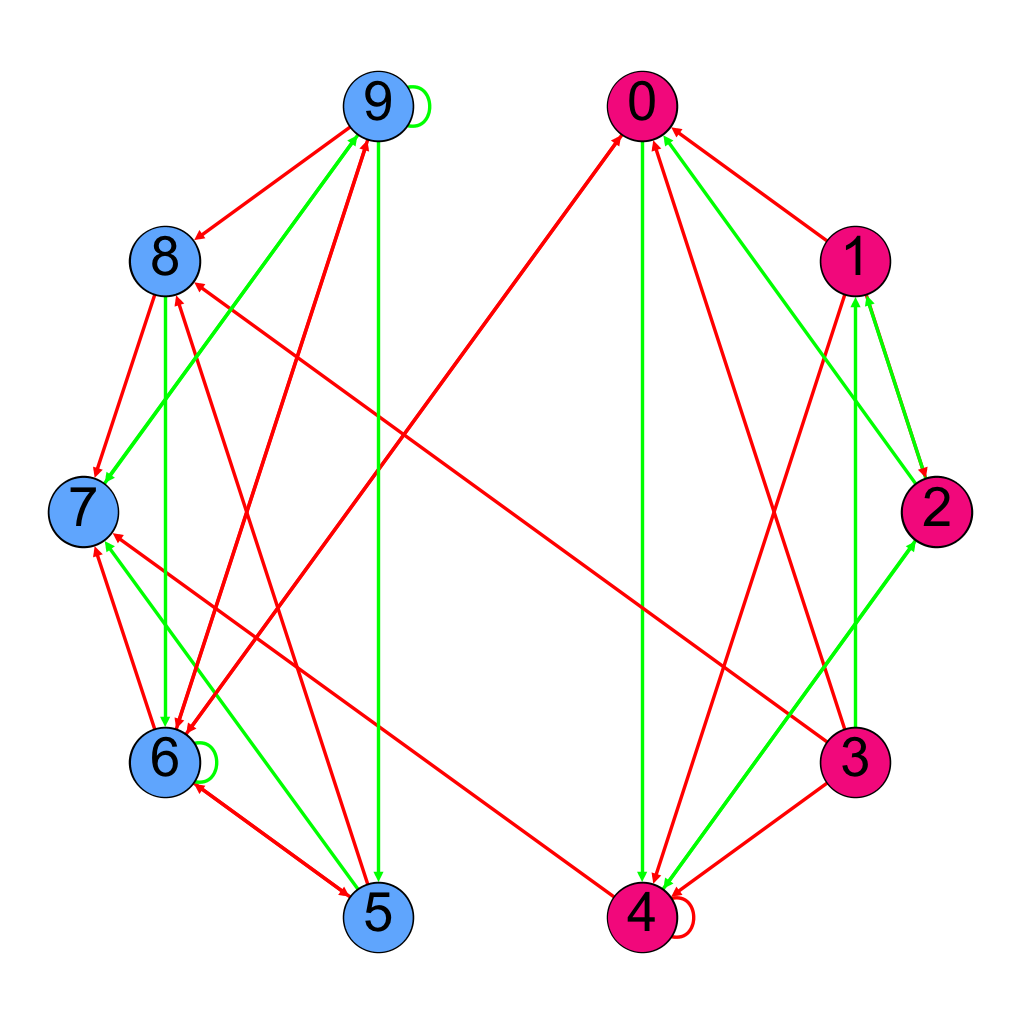}
		\caption{Parental network 2.}
	\end{subfigure}
	\begin{subfigure}[b]{0.4\linewidth}
		\includegraphics[width=\linewidth]{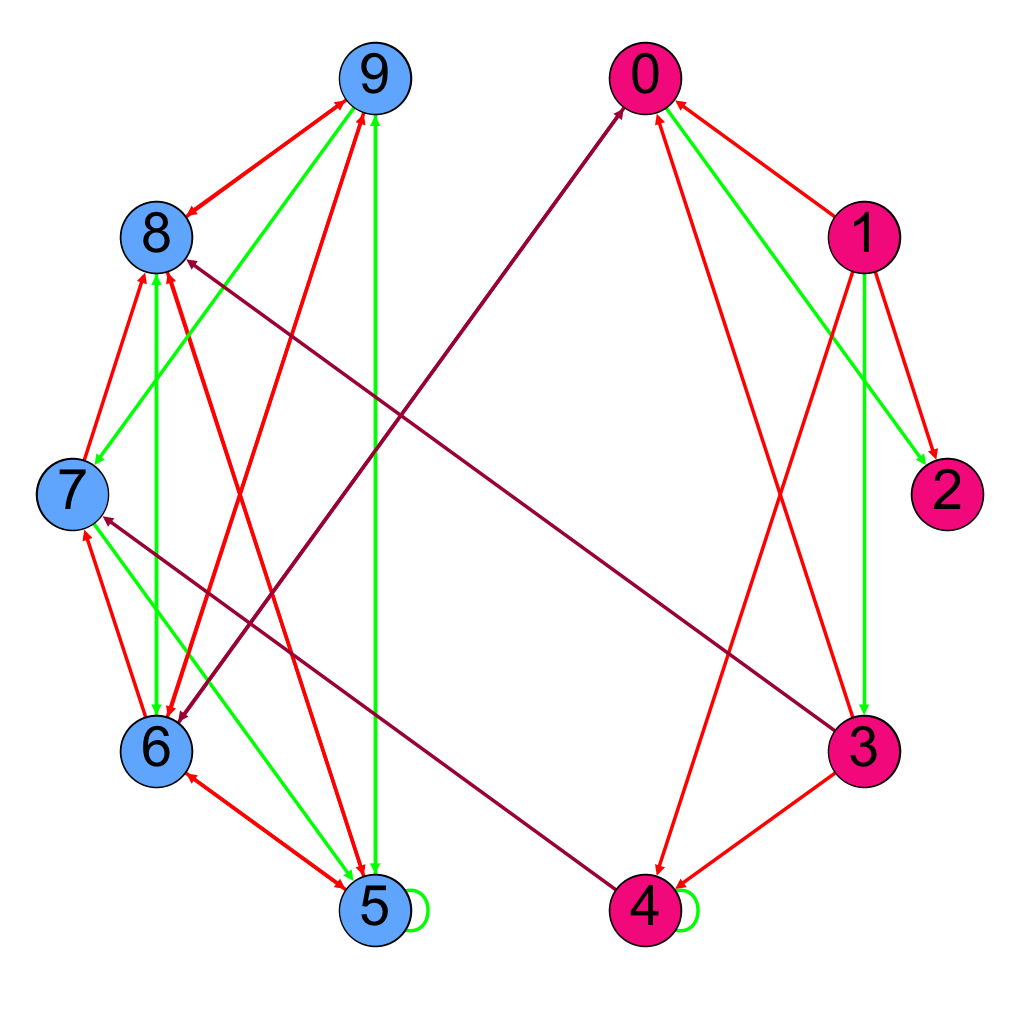}
		\caption{Child network 1.}
	\end{subfigure}
	\begin{subfigure}[b]{0.4\linewidth}
		\includegraphics[width=\linewidth]{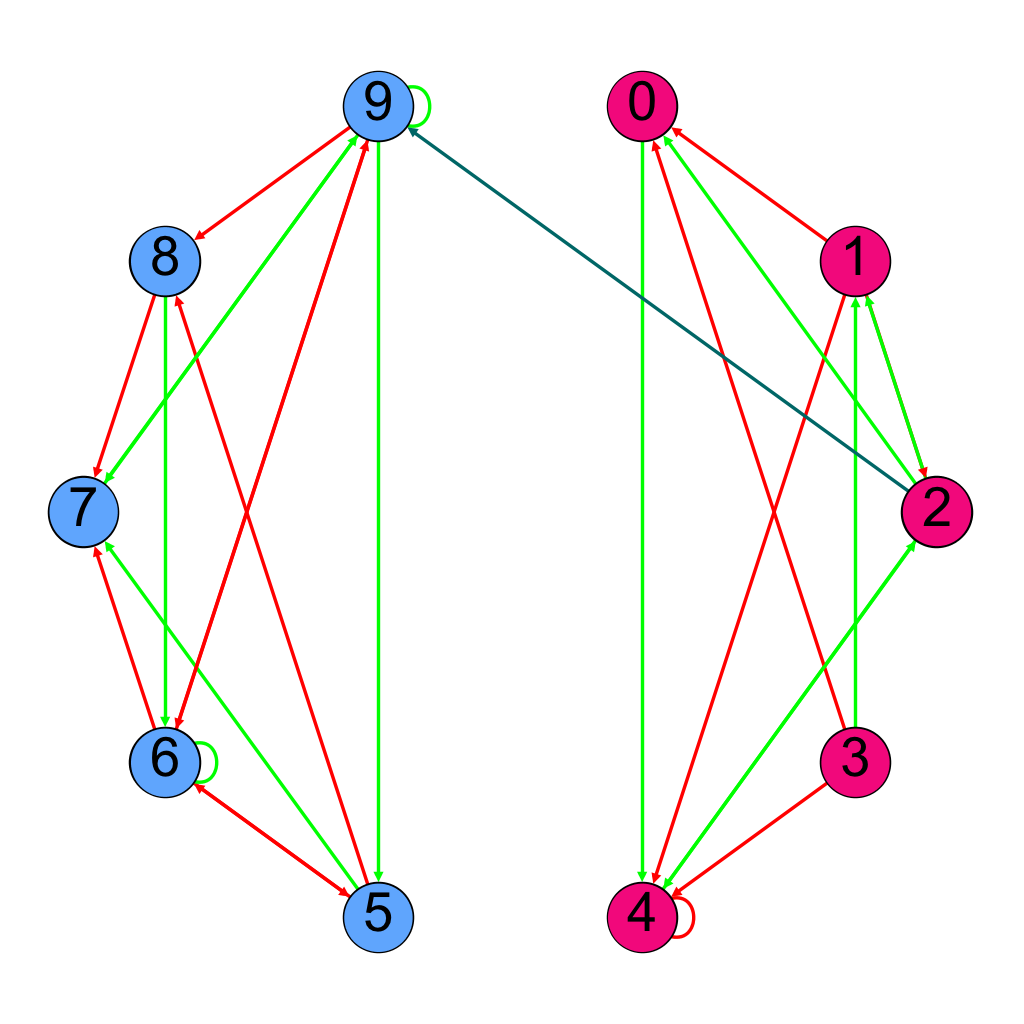}
		\caption{Child network 2.}
	\end{subfigure}
	\caption{Illustration of diagonal crossover}
	\label{fig:diagnonal-crossover}
\end{figure}

However this horizontal crossover may make the parental networks exchange not only modular clusters (defined by the nodes that follow a similar pattern in the activation targets~\cite{espinosa2010specialization,larson2016recombination}), but also some interactions between the two modules. This may corrupt modularity. In contrast, we use a crossover mechanism that swaps interactions between modules in a gene regulatory network with connections between modules in another network. We refer to this as diagonal crossover. Compared with horizontal crossover, this approach, as Figure \ref{fig:diagnonal-crossover} illustrates, should better preserve the community structure, since this approach will lead to higher eventual modularity Q score (Wilcoxon signed-rank test; $p<0.0019$).

\subsection{Modularity Metric}
We adopted the Q scoring system to quantify modularity in a network, based on the algorithm proposed by Newman~\cite{newman2004finding}. Briefly, this approach is defined as the difference between the ratio of the number of edges in the network connecting nodes within a module over the number of all the edges, and the same quantity when assigning the nodes into the same modules yet edges are assumed to be randomly connected in the network~\cite{kashtan2005spontaneous}. Formally, $Q$ is calculated as 
\begin{equation}
Q = \sum_{i}^{K}[\frac{l_i}{L} - (\frac{d_i}{2L})^2]
\end{equation}
where $i$ represents one of the $K$ potential modules within a network, $L$ is the total number of connections in a network, $l_i$ stands for the number of interactions in the module $i$, and $d_i$ is the sum of degrees of all the nodes in module $i$~\cite{espinosa2010specialization}. In other words, $Q$ considers the two ratios of both intra-module connection density and inter-module connection density~\cite{newman2004finding}. For a network to be considered high modularity, it must consist of as many within-module edges and as few inter-module edges as possible. Conversely, it will result in $Q=0$ if all the nodes are partitioned into the same module. 

The value $Q$ will sit in the range of $\left.[-\frac{1}{2}, 1\right.)$. Nodes in the gene regulatory network are partitioned into different groups according to their regulating gene activity patterns.

\section{Preliminary Experiments: Modularity Surprises}
\begin{figure}[htb]
	\centering
	\includegraphics[width=\linewidth]{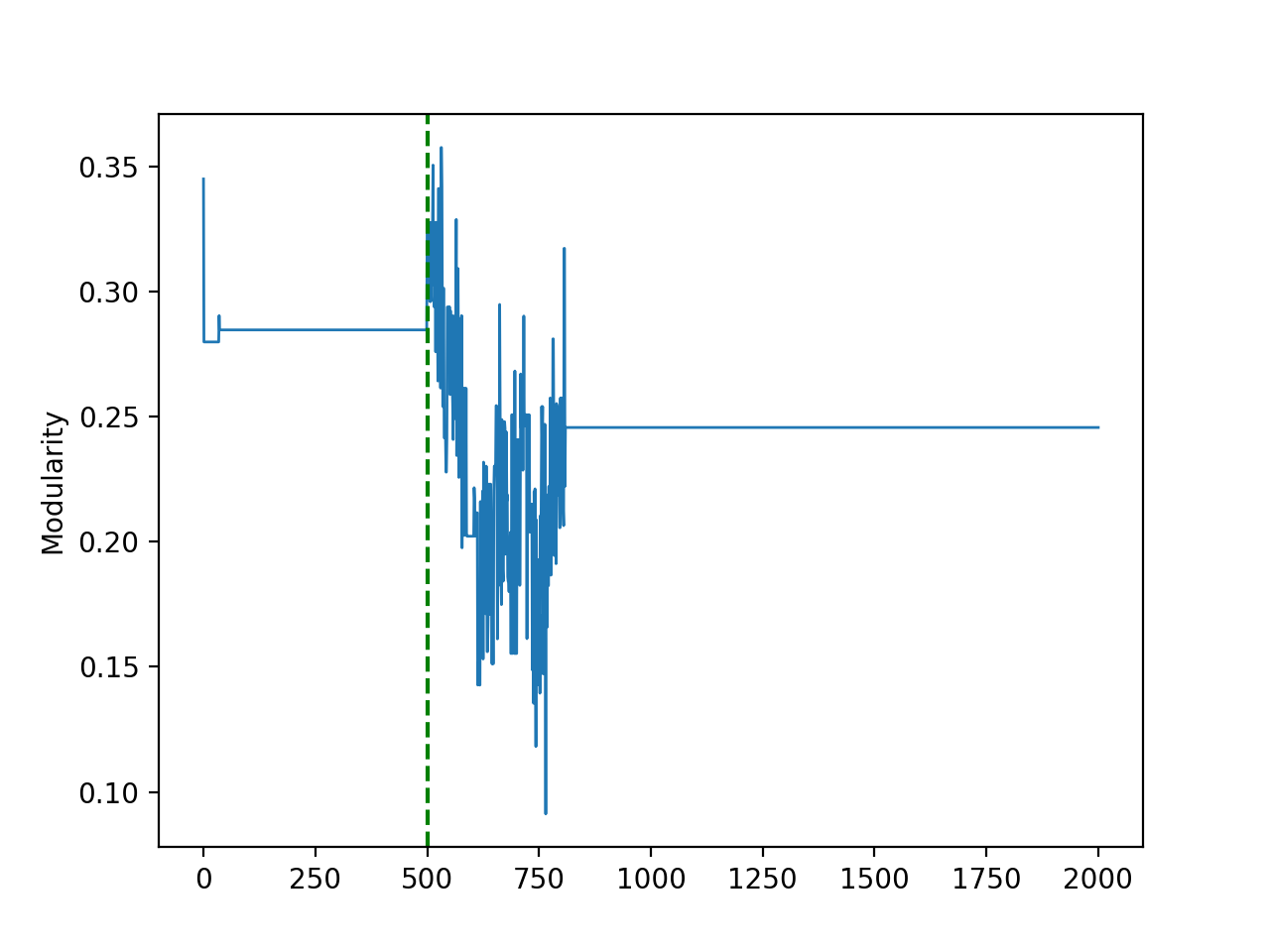}
	\caption{Modularity decreased after target change (marked by the vertical line).}
	\label{fig:early-modularity-not-work}
\end{figure}

Our baseline-setting experiments, using a standard genetic algorithm with an elite of 10 and tournaments of size 3, revealed surprising differences in the emergence of modularity from the results of ES\&W and Larson et al. Recall that in their experiments, overall modularity increased after a second target was added to the fitness function. In our initial experiments (using the Louvain metric~\cite{blondel2008fast} rather than the Q metric we use elsewhere), we instead observed a decrease immediately following the addition of the second target, with the overall modularity eventually stabilizing to a level substantially below that of the first phase (see Figure \ref{fig:early-modularity-not-work}).

Our settings differed from those of ES\&W and Larson et al. in the following ways:
\begin{enumerate}
\item{\label{crossuse}Use of crossover (difference from ES\&W only)}
\item{\label{tournuse}Tournament instead of proportional selection}
\item{\label{elituse}Incorporation of elitism}
\item{\label{paretouse}Omission of the age--fitness Pareto mechanism (difference from Larson et al. only,~\cite{Bongard2017})}
\end{enumerate}

Of these differences, item~\ref{crossuse} seems unlikely to explain our result since Larson et al. also used crossover, while items~\ref{elituse} and~\ref{paretouse} both increase the relative eagerness of our search. Item~\ref{tournuse} is more complex, since a tournament of size 3 exerts relatively weak selection pressure, but the relative pressure of tournament and proportional selection varies with the stage of the algorithm. Proportional selection depends on relative differences in fitness, so it typically exerts fairly strong pressure in early stages of search, but as the population fitness converges and differences reduce, pressure weakens; by contrast, tournament selection, being dependent only on fitness rank order, exerts a relatively constant selection pressure throughout. In particular, when populations are relatively converged (as at the time of the target switch), we would expect even relatively small tournaments to be more selective than proportional.

\begin{figure}[htb]
	\centering
	\includegraphics[width=\linewidth]{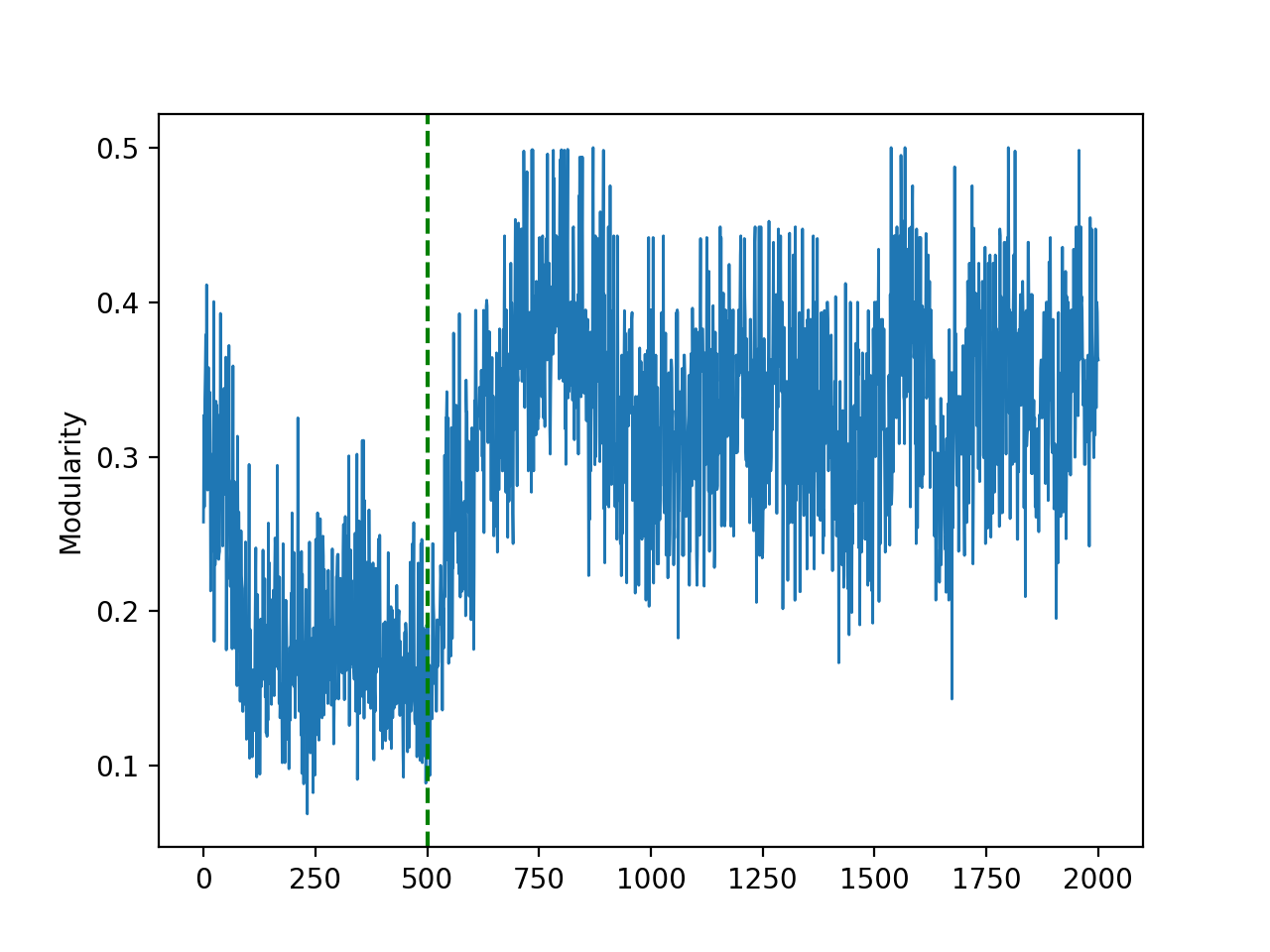}
	\caption{Without elitism, modularity increased after target change (marked by the vertical line).}
	\label{fig:modularity-worked-without-elitism}
\end{figure}

Based on these considerations, we decided to test the joint effects of elitism and tournament selection. The results bore out this hypothesis: the same algorithm and settings, with elitism eliminated and proportional substituted for tournament selection, led to the emergence of modularity, see Figure~\ref{fig:modularity-worked-without-elitism}. 

\comm{We applied the tournament selection scheme with the tournament size being three and the elitism mechanism with ten elites in every generation. 
As a result of this setting, the partition of the gene regulatory networks by the Louvain heuristics demonstrated a very low modularity score at the end of the simulated evolution. 
As  indicates, where the green line represents the generation to introduce specialization, by simulating the work in~\cite{espinosa2010specialization}, we had expected there would be a spike on modularity after gene specialization. 
Nevertheless, we observed a modularity decrease as a result after gene specialization.}

\comm{In order to understand this puzzling phenomenon, we removed the elitism mechanism and changed the tournament to proportional selection scheme. In consequence, we eliminated the deviant phenomenon as Figure \ref{fig:modularity-worked-without-elitism} indicates. Therefore, we hypothesized that the elitism mechanism or the tournament selection scheme hamper the evolutionary process on evolving out modular structures. }

\section{Experiment Settings}
\comm{Zhenyue, labels should always be semantically, not layout, based. You may want to change the grouping or ordering of sections, or of tables within sections. If you use a positional label, and the layout changes, this can quickly become impossibly confusing. 
Unless you have an extremely good reason, you should always use htb positioning, as latex may then be able to optimize space better. h-only positioning should only be used if you find the htb layout is absolutely disastrous.}

\begin{table}[htb]
	\centering
	\caption{Gene Activity Patterns}
	\label{table:genactiv}1
	\begin{tabu} to \linewidth {XX}  
		\toprule
		Target Pattern & Generation where \newline Pattern is added \\ 
		\midrule
		+1 -1 +1 -1 +1 -1 +1 -1 +1 -1 & 0 \\ 
		\midrule
		+1 -1 +1 -1 +1 +1 -1 +1 -1 +1 & 500 \\
		\bottomrule
	\end{tabu}
\end{table}

\begin{table}[htb]
	\centering
	\caption{Parameters of the Evolutionary Simulations}
	\label{table:simparams}
	\begin{tabu} to \linewidth {XXX}  
		\toprule
		Edge Size & Number of\newline Perturbations & Per-location\newline Perturbation Rate \\ 
		\midrule
		20 & 75 & 0.15 \\
		\bottomrule \toprule
		Mutation Rate & Population Size & Selection Type \\
		\midrule
		0.05 & 100 & Proportional \\
		\bottomrule \toprule
		Reproduction Rate & Maximum\newline Generation & Elite Size \\
		\midrule
		0.9 & 2000 & 0 or 10 \\
		\bottomrule \toprule
		Trials per\newline Treatment & Significance Test &  \\
		\midrule
		40 & Wilcoxon\newline Signed Rank & \\
		\bottomrule
	\end{tabu}
\end{table}

\begin{table}[htb]
	\centering
	\caption{Explanations of simulation parameters}
	\label{table:parmexplain}
	\begin{tabu} to \linewidth {X[1.3]X[3]} 
		\toprule
		Target\newline Patterns & patterns that are perturbed, and towards which gene regulatory networks evolve \\
		\midrule
		Target Addition\newline Generations & the generations where new targets are introduced \\
		\midrule
		Edge Size & the initial number of edges in the gene regulatory network \\
		\midrule
		Perturbation\newline Number & the number of perturbed versions of each gene activity pattern \\
		\midrule
		Perturbation Rate & the expected proportion of corrupted genes in a pattern \\
		\midrule
		Mutation Rate & the probability of a GRN node to gain or lose an interaction \\
		\midrule
		Population Size & the number of individuals in the population\\
		\midrule
		Selection Type & the type of selection used, and where tournament, the size of the tournament \\
		\midrule
		Reproduction Rate & the proportion of the population reproduced without change, vacancies being filled by the selection mechanism\\
		\midrule
		Maximum generation & the generation when the simulation will terminate  \\
		\bottomrule
	\end{tabu}
\end{table}

Tables~\ref{table:genactiv} and~\ref{table:simparams} show the gene activity patterns and the essential parameters of our evolutionary simulations. Unless otherwise specified, all experiments use the  stochastic fitness evaluation of ES\&W~\cite{espinosa2010specialization}. The only parameters that will vary from the tables are the selection type (tournament) and size, and the elite size.
The detailed explanations of these parameters are given in Table \ref{table:parmexplain}. 

The evaluation metrics for experiments include both the eventual fitness values and modularity Q scores from the last generation. All are significance tested using Wilcoxon's Signed-Rank Test.

\section{Experiments and Results}
\subsection{Diagonal Crossover Promotes Modularity}
\label{subsec:diag}

\begin{table}[htb]
	\centering
	\caption{Final Generation Best Fitness and Q Score with No, Horizontal and Diagonal Crossover}
	\label{table:crossfitq}
	\begin{tabu} to \linewidth {XXXX} 
		\toprule
		& No Crossover & Horizontal & Diagonal \\
		\midrule
		Fitness & 0.9476 & 0.9446 & 0.9488 \\ 
		\midrule
		Q Score & 0.1961 & 0.2919 & 0.3386 \\
		\bottomrule
	\end{tabu}
\end{table}

\begin{table}[htb]
	\centering
	\caption{Wilcoxon Ranked Sign Values for Table~\ref{table:crossfitq}}
	\label{table:crossfitqsig}
	\begin{tabu} to \linewidth {X[4]X[2.5]X[2.5]}  
		\toprule
		& Fitness P & Q Score P \\
		\midrule
		No < Horizontal & (Horizontal<No) 0.0882 & 1.7090e-6 \\ 
		\midrule
		Horizontal < Diagonal & 0.0006 & 0.0019 \\
		\bottomrule
	\end{tabu}
\end{table}
 
We ran trials comparing horizontal, diagonal and no crossover, in all cases without elitism.
As Tables \ref{table:crossfitq} and \ref{table:crossfitqsig} show, diagonal crossover generated significantly higher fitness and Q score than horizontal crossover, which in turn generated significantly higher Q score, though non-significantly lower fitness, than absence of crossover.

This Boolean model was proposed by Wagner~\cite{wagner1996does}, who showed that random recombination made no difference to evolution of stability. Our experiments suggest that more structured forms of recombination (which occur in biology) can contribute to evolvability. Diagonal crossover can preserve underlying network modules. Although horizontal crossover did not preserve community structures as well as diagonal, its partitioning is still based on a modular structure, and thus partially preserves modularity.
\comm{This can be the reason why both of these two crossover mechanisms could help in obtaining modularity, with diagonal crossover better than horizontal crossover. Meanwhile, different combinations of parental traits can increase the diversity of the population so that the evolution can be more exploratory.}

\subsection{Greedier Search Hampers Modularity}
\subsubsection{Elitism Hampers Modularity}~\\
\begin{table}[htb]
	\centering
	\caption{Final Generation Best Fitness and Q Score with and without Elites}
	\label{table:elitefitq}
	\begin{tabu} to \linewidth {XXX}
		\toprule
		& Without Elites & With 10 Elites \\
		\midrule
		Fitness & 0.9488 & 0.9472 \\ 
		\midrule
		Q Score & 0.3386 & 0.2735 \\
		\bottomrule
	\end{tabu}
\end{table}

\begin{table}[htb]
	\centering
	\caption{Wilcoxon Ranked Sign Values for Table~\ref{table:elitefitq}}
	\label{table:elitefitqsig}
	\begin{tabu} to \linewidth {X[5]X[2]X[2]}
		\toprule
		& Fitness P & Q Score P \\
		\midrule
		With 10 Elites < Without Elites & 0.0003 & 0.0022 \\ 
		\bottomrule
	\end{tabu}
\end{table}

We trialled simulations with an elite of 10 against simulations with no elites. Significantly lower best fitness (regulatory capability) and modularity arose when an elite was used (Tables~\ref{table:elitefitq} and \ref{table:elitefitqsig}).

\subsubsection{Proportional Selection generates Better Fitness than Tournament Selection, but Lower Modularity than Small Tournaments}~\\

\begin{table}[htb]
	\centering
	\caption{Final Generation Best Fitness and Q Score for Proportional Selection and Different Sized Tournaments}
	\label{table:selfitq}
	\begin{tabu} to \linewidth {XXXXX} 
		\toprule
		& Proport & Tourn\newline Size 2 & Tourn\newline Size 3 & Tourn\newline Size 10  \\
		\midrule
		Fitness & 0.9488 & 0.9404 & 0.9404 & 0.9371  \\ 
		\midrule
		Q Score & 0.3386 & 0.3697 & 0.3623 & 0.2783  \\
		\bottomrule
	\end{tabu}
\end{table}

\begin{table}[htb]
	\centering
	\caption{Wilcoxon Ranked Sign Values for Table~\ref{table:selfitq}}
	\label{table:selfitqsig}
	\begin{tabu} to \linewidth {X[5]X[2]X[2]}
		\toprule
		& Fitness P & Q Score P \\
		\midrule
		Proportional > Tourn Size 2 & 0.7401 & 0.0467 \\
		\midrule
		Tourn Size 3 < Size 2 & 0.9313 & 0.7881 \\ 
		\midrule
		Tourn Size 10 < Size 3 & 0.0164 & 0.0015 \\ 
		\midrule
		Tourn Size 10 < Proportional & 0.0227 & 0.0054 \\
		\bottomrule
	\end{tabu}
\end{table}
At least when populations are approaching convergence, tournament selection, especially with larger tournaments, imposes stronger selection pressure than proportional selection. In these results, we see in Tables~\ref{table:selfitq} and~\ref{table:selfitqsig} fairly much the anticipated decline in ultimate fitness with increasing selection pressure (though none of the differences are significant at the 1\% level). Interestingly, proportional selection exhibits an unexpected (but non-significant) lower modularity than tournaments of size 2 or 3, but generates significantly more modularity than tournaments of size 10.


\subsection{Dynamically Stochastic Fitness Evaluation generates Higher Fitness and Modularity than Static}

\begin{table}[htb]
	\centering
	\caption{Final Generation Best Fitness and Q Score for Dynamic and Static Stochastic Fitness Evaluation}
	\label{table:dynfitq}
	\begin{tabu} to \linewidth {XXX}
		\toprule
		& Dynamic & Static \\
		\midrule
		Fitness & 0.9488 & 0.9379 \\ 
		\midrule
		Q Score & 0.3386 & 0.2948 \\
		\bottomrule
	\end{tabu}
\end{table}

\begin{table}[htb]
	\centering
	\caption{Wilcoxon Ranked Sign Values for Table~\ref{table:dynfitq}}
	\label{table:dynfitqsig}
	\begin{tabu} to \linewidth {X[5]X[2]X[2]}
		\toprule
		& Fitness P & Q Score P \\
		\midrule
		Dynamic < Static & 3.5669e-8 & 0.0167 \\ 
		\bottomrule
	\end{tabu}
\end{table}

We compared dynamically stochastic fitness evaluation (in which the perturbations are generated anew each generation) and static fitness evaluation (the perturbations are generated once for all at the start of each run). Dynamic fitness evaluation outcompeted static on both survivability and modularity (Tables~\ref{table:dynfitq} and \ref{table:dynfitqsig}), though the latter result is not significant at the 1\% level.

\section{Analysis and Discussion}
\subsection{Modular systems did not gain dominance via selection for robust target recovery}
\label{subsec:nondom}
While some algorithm variants permitted modularity to emerge and survive under selection for robust target recovery, it did not (in extreme contrast to natural systems~\cite{schlosser2004modularity}) come to dominate populations. We saw some indication that greedier mechanisms, including elitism and tournament selection scheme, impeded the emergence of modularity. This implies that individuals who performed best, particularly in early stages of evolution, might not be especially modular. In general, the most competitive elites in each generation did not have the most modular gene regulatory networks. 

Overall, this suggests that while the ES\&W framework has been useful to demonstrate particular properties of the emergence of modularity, in particular from pressure on gene specialization, it may not be sufficiently nature-like to function as a useful testbed to explore algorithmic effects on modularity emergence. If we see differences in algorithm behavior on this testbed, we will not know whether they arise from the abstractions from the natural environment (in which case they may usefully explain the behavior of natural environments, and suggest how to extend that behavior to artificial systems), or whether they arise from the substantial differences between the abstraction and reality (in which case any extrapolations would be moot). 
\comm{the modularity emergence condition, namely gene specialization promotes modular networks, may not be plausible to explain biological modularity. They indicated that modules in the simulated gene regulatory networks did not gain dominance in determining the survivability of individuals. However, biologically, modular networks are dominant and ubiquitous~\cite{schlosser2004modularity}.} 

\begin{table}[htb]
	\centering
	\caption{Modularity dominance for data from subsection~\ref{subsec:diag}}
	\label{table:2000notmodularity}
	\begin{tabu} to \linewidth {XXX}  
		\toprule
		Generation Range & Modularity & Fitness \\
		\midrule
		(500, 2000) & 0.5000 & 0.9482 \\ 
		\midrule
		 & 0.1736 & 0.9502 \\
		\bottomrule
	\end{tabu}
\end{table}

To further investigate the behavior of this system, we took more detailed measurements from the  simulations in subsection~\ref{subsec:diag}, using the diagonal crossover. Specifically, from each run, we collected the fittest gene regulatory network among networks that were the most modular; and conversely, the network that was least modular among those that had the greatest fitness value. 
We expected the mean fitness of the latter to be lower than the former. Surprisingly, the situation was reversed: less modular networks generally recovered the target more robustly than more modular, as shown in Table~\ref{table:2000notmodularity}. This does not reflect biological observations.

\begin{table}[htb]
	\centering
	\caption{Modularity dominance for extended runs from more complex environments}
	\label{table:35000notmodularity}
	\begin{tabu} to \linewidth {XXX}
		\toprule
		Generation Range & Modularity & Fitness \\
		\midrule
		(26000, 35000) & 0.5506 & 0.9100 \\ 
		\midrule
		& 0.4151 & 0.9419 \\
		\bottomrule
	\end{tabu}
\end{table}
We wondered whether this inconsistency could arise from insufficient complexity in the targeted gene activity patterns. The number of genes in patterns might be too simple, or the number of targets might be too few, to reflect natural environments. Perhaps modular network might give great performance on complex tasks, but worse than non-modular ones for simple tasks. Thus using the basic set-up of subsection~\ref{subsec:diag}, we ran more complicated evolutionary simulations
using patterns comprising 15 nodes, encountering a sequence of seven different targets. Evolution was extended to 35,000 generations and during the final epoch from $(26000 \rightarrow 35000)$ generations, it was evolving to robustly recover all seven targets. We repeated the preceding analysis; the results in Table~\ref{table:35000notmodularity} resemble those of Table~\ref{table:2000notmodularity}. Overall, the number and complexity of targets could not resolve the issue: less modular networks still recovered the target more robustly than more modular networks. 

\subsection{Inter-Module Connections Can Hamper Network Fitness}
\begin{figure}[htb]
	\centering
	\begin{subfigure}[b]{0.45\linewidth}
		\includegraphics[width=\linewidth]{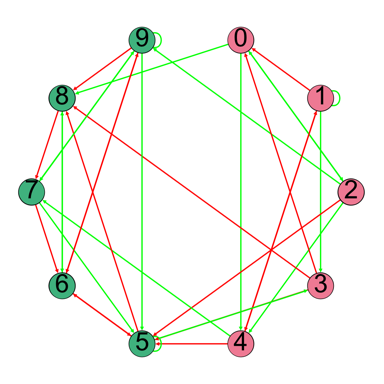}
		\caption{Before removal}
	\end{subfigure}
	\begin{subfigure}[b]{0.45\linewidth}
		\includegraphics[width=\linewidth]{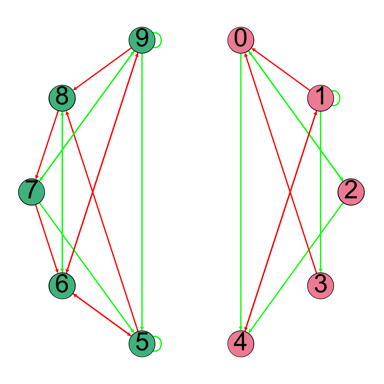}
		\caption{After removal}
	\end{subfigure}
	\caption{Illustration of inter-module connection removal.}
	\label{fig:connection-removal}
\end{figure}
To further investigate this phenomenon, we took the 40 networks that were least modular among those having the greatest fitness value from the first experiment in subsection~\ref{subsec:nondom}. We wondered what would happen if we simply removed all non-modular interconnections. So we did so with all 40 networks, and measured fitness after this removal (modularity was, of course, perfect).
Of these 40 relatively low-modularity (Q) but near-optimal fitness networks, 24 exhibited \emph{even higher} fitness after manually deleting inter-module edges. That is, more than half these originally non-modular networks exhibited better fitness performance after removing all the inter-module connections. For example, the right network in Figure \ref{fig:connection-removal} was the consequence of removing inter-module connections of the network in the left. Its fitness value of the latter was 0.9502, a result of removing~20\% of the connections from a network whose fitness was only 0.9472. 
So these 24 much-improved networks (higher fitness than any found in the run, but also much higher modularity) were not only available to the evolutionary algorithm -- they were even relatively nearby (a few edge deletion mutations away). Yet the algorithm reliably did not find them!

Originally, we suspected this anomaly might result from their lower edge density -- perhaps they were too much below the edge density targeted by the biased mutation operator, so that this soft constraint eliminated them from the search. Further investigation revealed that on average, they still retained approximately 30 edges, which is above the targeted density of the mutation operator (see subsection~\ref{subsec:bias}), so far from being difficult to reach, the biased mutation operator favored moving toward them.

\begin{figure}[htb]
	\centering
	\includegraphics[width=\linewidth]{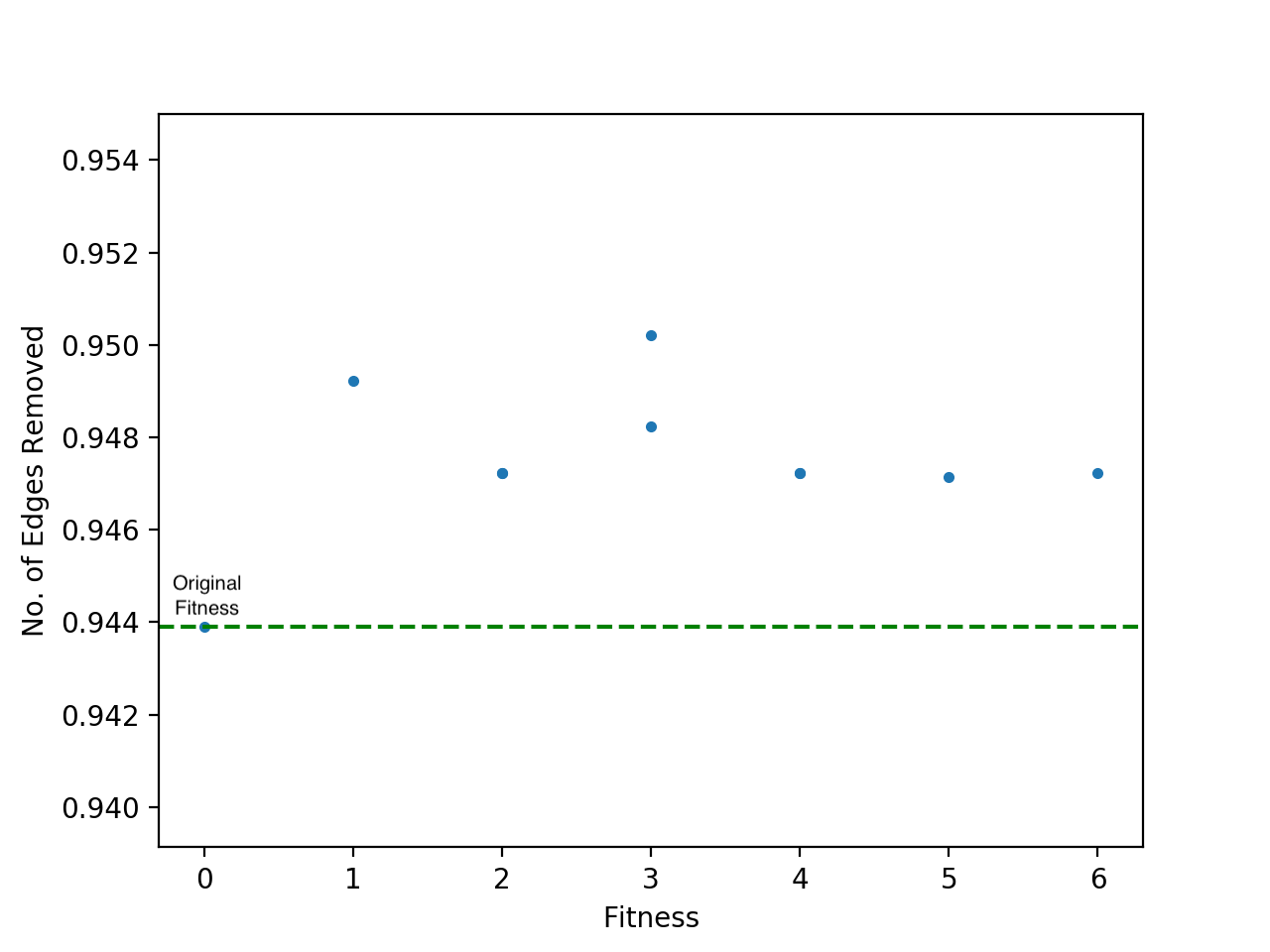}
	\caption{Fitness Values along all Inter-module Edge Removal Paths from a High Fitness, Low Modularity Network that results in Increased Final Fitness.}
	\label{fig:soto-path}
\end{figure}
In order to further comprehend this phenomenon, that our evolutionary simulations could not find a path to the trimmed networks, we recorded all fitness values that could be obtained by removing one inter-module edge in turn, until all have been deleted, and plotted them as graphs. Figure~\ref{fig:soto-path} is typical. We could usually see a steadily improving fitness as edges were deleted, along a path that was favored by the biased mutation operator, yet our genetic algorithm could not find these paths.

\subsection{Temporally fluctuating  landscapes seem essential for generating modularity with a standard genetic algorithm}
Because ES\&W~\cite{espinosa2010specialization} re-sampled the perturbations used for fitness evaluation each generation, their fitness values varied dynamically and stochastically. By contrast, Larson et al. generated their samples once for each run, so that the fitness, while still stochastic, was not dynamic. While they saw useful emergence of modularity with their age-Pareto algorithm, which certainly generates higher population diversity and greater exploration, we did not have access to such an algorithm during the research on this paper. In any case, our purpose was to gain understanding in the context of the immense research knowledge that has been accumulated about standard genetic algorithms. Additionally, to increase the emphasis on modularity, we made the crossover always occur at the boundary of the two modules. That is, crossover would always interchange inter-module connections. We did not see modularity emerge with a standard genetic algorithm on a static landscape. 

\begin{table}[htb]
	\centering
	\caption{Fitness and Q Scores for Neighbors of Final Generation Fittest Individuals in Static and Dynamic Environments}
	\label{table:dynstat}
	\begin{tabu} to \linewidth {XXX} 
		\toprule
		& Dynamic & Static \\
		\midrule
		Original Fitness & 0.9461 & 0.9323 \\ 
		\midrule
		Best Neighbor Fitness & 0.9410 & 0.9323 \\
		\midrule
		Original Q & 0.3374 & 0.1851 \\
		\midrule
		Best Neighbor Q & 0.3791 & 0.2223  \\
		\bottomrule
		
	\end{tabu}
\end{table}
\comm{\begin{table}[htb]
	\centering
	\caption{Wilcoxon Ranked Sign Values for Table~\ref{table:dynstat}}
	\label{table:dynstatsig}
	\begin{tabu} to \linewidth {X[4]X[2.5]X[2.5]} 
		\toprule
		& Fitness P & Q Score P \\
		\midrule
		Original Static < Dyna & 0.7223 & 2.6879e-5 \\ 
		\midrule
		Neighbor Static < Dyna & 0.9826 & 2.2956e-5 \\ 
		\bottomrule
	\end{tabu}
\end{table}}

To further understand this phenomenon, we collected the gene regulatory networks of the final generation, and mutated each network 499 times to generate neighbors. \comm{That is, each network would have 500 neighbors, given including itself.} We measured the fitness values of these neighbors with the target perturbations from this generation, and determined their maximum. In this fashion, we would have 40 neighborhood-maximum fitness values for both dynamic and static fitness evaluation. We repeated this process for the modularity Q score. \comm{Formally, a maximum value for a network $N$ is collected with the formula 
\begin{equation}
max(function(mutatedneighbours(N)))
\end{equation}
where $function$ can either be $fitness$ or $modularity$. Subsequently, our statistical test indicated that fitness of stochastic neighbors did not demonstrate advantages, whereas their modularity Q scores were much higher than deterministic neighbors, as}
As Table~\ref{table:dynstat} shows, the fittest individuals in the final generation for the dynamic problem were generally local optima, whereas for the static problem, they were generally on a fitness plateau, with equally fit neighbors, and a substantially lower fitness than found in the dynamic problem. In both scenarios, there were neighbors of substantially higher modularity than the original individual, but overall modularity, both of the final generation best individual and of its neighbors, were much higher in the dynamic problem than in the static.
\comm{In general, in order to evolve out high modularity, a combination of gene specialization and a constantly changing environments will be desirable, instead of applying gene specialization alone.

Moreover, previously the statistics test revealed that the stochastic approach would lead to a higher fitness value, whereas this advantage disappeared when evaluating the fitness of mutated neighbors. Further investigation suggested that a deterministic, or static landscape may result in the searching getting stuck at the local optima. This is because for our 40 networks generated by deterministic fitness evaluation, the maximum fitness values for a network's neighbors were all from itself. That is, the neighbors of a network evolving in a static landscape always performed worse than themselves. Formally, for a network $N$, 
\begin{equation}
max(fitness(mutatedneighbours(N))) = fitness(N)
\end{equation}
Furthermore, there existed a lot of networks produced by deterministic fitness evaluation whose fitness values were much lower (approximately 0.88), compared to the rest of networks as well as those generated by stochastic fitness evaluation (approximately 0.93). We hypothesized that these low-performing networks are the cause on why statistically, fitness values generated by deterministic evaluation were lower than stochastic evaluation. Additionally, there may exist some correlation between getting stuck at local optima and modularity evolution. }

\comm{
\subsection{More modular networks require fewer connections}
As previous results suggested, interactions between modules sometimes do not contribute to and even hamper the regulation activity of networks. That is, a network can gain a better performance by removing those inter-module connections, which indicates that modular networks require fewer connections in total. In order to justify this hypothesis, we collected both of the most and the least modular network among those fittest individuals from each evolutionary simulation in Section 4.1, using the diagonal crossover. That is, given two networks that have the same fitness value, we would like to discover whether the more modular one needs fewer connections. Our statistical test verified this hypothesis to be correct, as Table \ref{table:4.14} indicates. 
\begin{table}[htb]
	\centering
	\caption{Results for verifying more modular networks require fewer connections}
	\label{table:4.14}
	\begin{tabular}{| p{0.225\linewidth}  | p{0.225\linewidth}  | p{0.225\linewidth} | p{0.225\linewidth} |} 
		\hline
		& Most Modular & Least Modular & Most < Least Modular p\\
		\hline
		Edge Number & 24.6 & 29.925 & 6.1913e-7\\ 
		\hline
	\end{tabular}
\end{table}

Clune et al. stated that the evolutionary origin of modularity is due to the cost associated with every connection in the network~\cite{clune2013evolutionary}. They demonstrated this by their experiments indicating that there was a significant emergence of modular networks after imposing a penalty on the number of edges in the network~\cite{clune2013evolutionary}.That is, modularity arose in order to minimise the connection costs. Specifically, they made simulated organisms evolve towards two objectives, namely to maximize the performance and to minimise the edge costs. However, in reality, biological organisms evolve in a single-objective fashion. That is, they are only selected under the pressure of fitting the living environments. Therefore, the theory stating that modularity comes from minimizing connection costs may not be sufficiently plausible. 

Our results revealed a converse causality of Clune et al.'s explanation on modularity. To be specific, the connecting costs of modular networks are lower may be because modular networks need fewer edges to support their activities than non-modular ones. It may be also due to this, Clune et al. can recognise and select more modular systems by choosing structures in which there are fewer connections. Nevertheless, containing fewer edges is a property of more modular networks, not their evolutionary origin. 
}

\section{Conclusions}
It is of considerable importance to identify those characteristics of evolutionary algorithms that will lead to modular problem solutions. Our belief that this is generally feasible is heavily reliant on the ubiquity of modular solutions in biological evolution. So useful testbeds for exploring algorithms' propensity to generate modular solutions need to abstract the relevant aspects of the real world. We have seen that from this perspective, ES\&W's environment exhibits a number of anomalies. It is highly sensitive to the eagerness of search (suggesting that differences observed in the behavior of different algorithms might be ascribed to the propensity to generate modularity, while in fact merely reflecting algorithm eagerness). While it supports the emergence of modularity, it does not appear to support their dominance, in contrast to real-world behavior. The non-dominance of modularity is surprising. We can manually find high-fitness paths that lead to modular solutions of even higher fitness, and these paths are favored by the mutation bias, yet the algorithm does not find them. This appears to be a result of the dynamic nature of the fitness function: the paths to these modular solutions may not appear favorable to the algorithm because of the dynamic stochastic variations in the fitness landscape. Paradoxically, this dynamic aspect seems essential to the emergence of modularity at all. The static variant of this problem does not support the emergence of modularity under standard genetic algorithms (we accept that it does so under strongly diversity-encouraging algorithms such as age-Pareto algorithms). This appears to be linked to the genetic algorithms becoming trapped in flat, featureless regions of the fitness landscape, which the small fluctuations in the fitness landscape of the dynamic variant allow them to escape.

These surprising properties of the problem render it problematic as a benchmark test. While we have gained some understanding of the problem domain, it is clear that deeper understanding is necessary, if we are to either rehabilitate it as a useful benchmark, or to find variants that may not behave so paradoxically. We are pursuing further investigations into the properties of the problem domain.

\comm{In summary, we found that the diagonal crossover mechanism can promote the emergence of modularity. In contrast, elitism hampers the rise of modular networks, which indicates that early optimal individuals did not demonstrate high-level modularity. Further experiments also indicated that the theory on the origin of modularity resulting from specialization has limitations on explaining the surviving dominance of modular systems in biology. Furthermore, networks that have high fitness values could demonstrate better performance after converting them into modular structures by removing their inter-module edges. This suggests that modular networks initialized by gene specialization may evolve towards structures requiring a fewer number of connections in total. However, evolutionary simulations could not find these more optimal solutions. Moreover, individuals that live in more fluctuant environments can result in more modular network structures. Therefore, fluctuant landscapes can be essential for modularity evolution. In the future, we will aim to understand the reason why genetic algorithms could not find a path to individuals with better fitness and modularity. Additionally, we would also investigate the correlation between fluctuation of landscapes and the level of modularity. }

\comm{\section*{Acknowledgements}
?Do we need to acknowledge grants here? Assistance from Bongard? etc.}

\bibliographystyle{ACM-Reference-Format}
\bibliography{bib/zhenyue,bib/tom,bib/bob}

\end{document}